\documentclass[runningheads]{llncs}
\usepackage{amsmath}
\usepackage{amssymb}

\usepackage{graphicx}
\usepackage{cite}

\usepackage{color} 
\usepackage{epsfig}
\usepackage{amsfonts}

\begin{document}
\title{Recognizing Facial Expressions of Occluded Faces using Convolutional Neural Networks\thanks{Research supported by Novustech Services through Project 115788 (Innovative Platform based on Virtual and Augmented Reality for Phobia Treatment) funded under the POC-46-2-2 by the European Union through FEDR.}}
%
%
\author{Mariana-Iuliana Georgescu\inst{1,2} \and
Radu Tudor Ionescu\inst{1}$^{,\diamond}$}
\titlerunning{Recognizing Facial Expressions of Occluded Faces using Neural Networks}
\authorrunning{Mariana-Iuliana Georgescu, Radu Tudor Ionescu}
%
\institute{
University of Bucharest, 14 Academiei Street, Bucharest, Romania\\
\and
Novustech Services, 12B Aleea Ilioara, Bucharest, Romania\\
$^{\diamond}$\email{raducu.ionescu@gmail.com}}
\maketitle              

\begin{abstract}
In this paper, we present an approach based on convolutional neural networks (CNNs) for facial expression recognition in a difficult setting with severe occlusions. More specifically, our task is to recognize the facial expression of a person wearing a virtual reality (VR) headset which essentially occludes the upper part of the face. In order to accurately train neural networks for this setting, in which faces are severely occluded, we modify the training examples by intentionally occluding the upper half of the face. This forces the neural networks to focus on the lower part of the face and to obtain better accuracy rates than models trained on the entire faces. Our empirical results on two benchmark data sets, FER+ and AffectNet, show that our CNN models' predictions on lower-half faces are up to $13\%$ higher than the baseline CNN models trained on entire faces, proving their suitability for the VR setting. Furthermore, our models' predictions on lower-half faces are no more than $10\%$ under the baseline models' predictions on full faces, proving that there are enough clues in the lower part of the face to accurately predict facial expressions. 
\keywords{Facial expression recognition \and convolutional neural networks  \and severe face occlusion \and virtual reality headset.}
\end{abstract}
\section{Introduction}

Facial expression recognition from images is an actively studied problem in computer vision, having a broad range of applications including human behavior understanding, detection of mental disorders, human-computer interaction, among others. Our particular application is to recognize the facial expressions of a person wearing a virtual reality (VR) headset and use the recognition result in order to provide feedback to the VR system, which can automatically change the VR experience according to the user's emotions.

\begin{figure}[!t]
\centering
\includegraphics[width=0.8\textwidth]{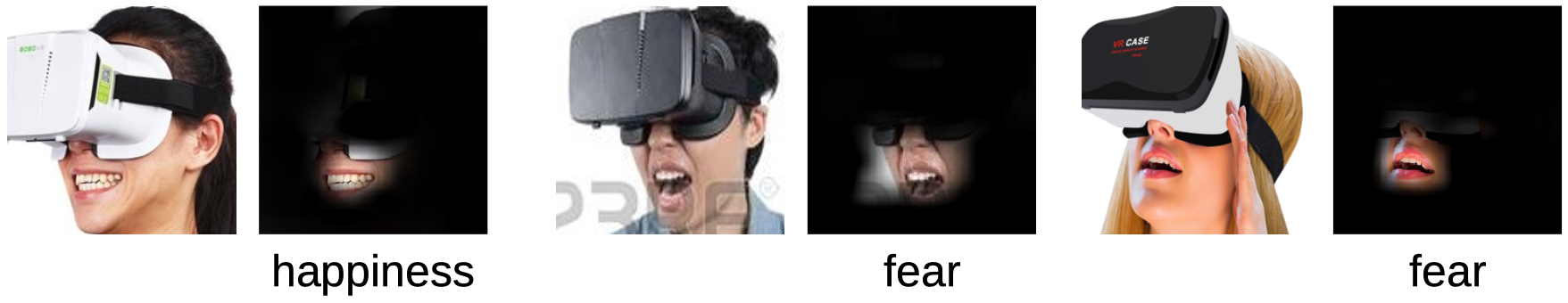}
\caption{Images (of people wearing VR headsets) with corresponding Grad-CAM~\cite{Selvaraju-ICCV-2017} explanation masks and labels from a VGG-face model trained on lower-half images.}\label{fig_visual_explain_vr}
\end{figure}

In the past few years, most works~\cite{Barsoum-ICMI-2016,Ding-FG-2017,Giannopoulos-2018,Hasani-CVPRW-2017,Hua-Access-2019,Kim-JMUI-2016,Li-MTA-2017,Li-CVPR-2017,Liu-CVPRW-2017,Li-ICPR-2018,Meng-FG-2017,Mollahosseini-CVPRW-2016,Tang-WREPL-2013,Wen-CC-2017,Yu-ICMI-2015,Zeng-ECCV-2018} have focused on building and training deep neural networks in order to achieve state-of-the-art results. Engineered models based on handcrafted features~\cite{Al-Chanti-VISIGRAPP-2018,Ionescu-WREPL-2013,Shah-PRL-2017,Shao-PRL-2015} have drawn very little attention, since such models usually yield less accurate results compared to deep learning models. As most recent works, we adopt deep convolutional neural networks (CNNs) due to their capability of attaining state-of-the-art results in facial expression recognition. However, we have to train the neural networks for a very difficult setting, in which the person to be analyzed wears a VR headset. The currently available VR headsets essentially occlude the upper part of the face, posing significant problems for a standard (not adapted) facial expression recognition system. Our goal is to adapt the facial expression recognition system for this specific setting. To achieve this goal, we train two CNN models, VGG-f~\cite{Chatfield-BMVC-14} and VGG-face~\cite{Parkhi-BMVC-2015}, on modified training images in which the upper half of the face is completely occluded. This forces the neural networks to find discriminative clues in the lower half of the face, as shown in Figure~\ref{fig_visual_explain_vr}. 

We perform experiments showing that our models (trained with occluded faces) obtain better accuracy rates than models trained on the entire faces, when the test set contains occluded faces. The experiments are conducted on two benchmark data sets, FER+~\cite{Barsoum-ICMI-2016} and AffectNet~\cite{Mollahosseini-TAC-2019}. Our empirical results show that our CNN models' trained on lower-half faces are up to $13\%$ higher than the baseline CNN models trained on entire faces, when the test set includes images of lower-half faces. Furthermore, our models trained and tested on lower-half faces are about $10\%$ (or even less) under the baseline models, when the baseline models are tested on full faces, thus proving that there are enough clues in the lower part of the face to accurately predict facial expressions. 
Overall, our empirical results indicate that learning and inferring facial expressions solely on the lower half of the face is a viable option for recognizing facial expression of persons wearing VR headsets. We thus conclude that our final goal, that of providing feedback to the VR system in order to automatically change the VR experience based on the user's emotions, is achievable.

We organize the rest of this paper as follows. We discuss related work in Section~\ref{sec_related_work}. We present the convolutional neural networks in Section~\ref{sec_method}. We describe the empirical results in Section~\ref{sec_experiments}. Finally, we draw our conclusions in Section~\ref{sec_conclusion}.

\section{Related Art}
\label{sec_related_work}

The early works on facial expression recognition are mostly based on handcrafted features~\cite{Tian-HFR-2011}. After the success of AlexNet~\cite{Hinton-NIPS-2012} in the ImageNet Large Scale Visual Recognition Challenge (ILSVRC)~\cite{Russakovsky2015}, deep learning has been widely adopted in the computer vision community. Perhaps some of the first works to propose deep learning approaches for facial expression recognition were presented at the 2013 Facial Expression Recognition (FER) Challenge~\cite{Goodfellow-ICONIP-2013}. Interestingly, the top scoring system in the 2013 FER Challenge is a deep convolutional neural network~\cite{Tang-WREPL-2013}, while the best handcrafted model ranked only in the fourth place~\cite{Ionescu-WREPL-2013}. With only a few exceptions~\cite{Al-Chanti-VISIGRAPP-2018,Shah-PRL-2017,Shao-PRL-2015}, most of the recent works on facial expression recognition are based on deep learning~\cite{Barsoum-ICMI-2016,Ding-FG-2017,Giannopoulos-2018,Georgescu-Access-2019,Hasani-CVPRW-2017,Hua-Access-2019,Kim-JMUI-2016,Li-MTA-2017,Li-CVPR-2017,Liu-CVPRW-2017,Li-ICPR-2018,Meng-FG-2017,Mollahosseini-CVPRW-2016,Wen-CC-2017,Yu-ICMI-2015,Zeng-ECCV-2018}. Some of these recent works~\cite{Hua-Access-2019,Kim-JMUI-2016,Li-MTA-2017,Wen-CC-2017,Yu-ICMI-2015} proposed to train an ensemble of convolutional neural networks for improved performance, while others~\cite{Connie-MIWAI-2017,Kaya-IVC-2017} combined deep features with handcrafted features such as SIFT~\cite{Lowe-SIFT-2004} or Histograms of Oriented Gradients (HOG)~\cite{Dalal-HOG-2005}. Works that combine deep and handcrafted features usually employ a single CNN model and various handcrafted features, e.g. Connie et al.~\cite{Connie-MIWAI-2017} employed SIFT and dense SIFT, while Kaya et al.~\cite{Kaya-IVC-2017} employed SIFT, HOG and Local Gabor Binary Patterns (LGBP). 
While most works studied facial expression recognition from static images as we do here, some works approached facial expression recognition in video~\cite{Hasani-CVPRW-2017,Kaya-IVC-2017}. 

Different from these mainstream works~\cite{Barsoum-ICMI-2016,Connie-MIWAI-2017,Ding-FG-2017,Giannopoulos-2018,Hasani-CVPRW-2017,Hua-Access-2019,Ionescu-WREPL-2013,Kaya-IVC-2017,Kim-JMUI-2016,Li-MTA-2017,Li-CVPR-2017,Liu-CVPRW-2017,Meng-FG-2017,Mollahosseini-CVPRW-2016,Shah-PRL-2017,Shao-PRL-2015,Wen-CC-2017,Yu-ICMI-2015,Zeng-ECCV-2018}, we focus on recognizing facial expressions of occluded faces. More closely related to our work, Li et al.~\cite{Li-ICPR-2018} proposed an end-to-end trainable Patch-Gated CNN that can automatically perceive occluded region of the face, making the recognition based on the visible regions. To find the visible regions of the face, their model decomposes an intermediate feature map into several patches according to the positions of related facial landmarks. Each patch is then reweighted by its importance, which is determined from the patch itself.  Different from Li et al.~\cite{Li-ICPR-2018}, we consider a more difficult setting in which half of the face is occluded. 

To our knowledge, the only work that studies facial expression recognition for people wearing VR headsets is that of Hickson et al.~\cite{Hickson-WACV-2019}. In their work, Hickson et al.~\cite{Hickson-WACV-2019} presented an algorithm that automatically infers expressions from images of the user's eyes captured from an infrared gaze-tracking camera mounted inside the VR headset, while the user is engaged in a VR experience. While their approach is applicable to VR headsets that have an infrared camera mounted inside, our approach, which uses an external camera, is cheaper and applicable to all VR headsets, thus being more generic. 

\section{Method}
\label{sec_method}

In this work, we choose two pre-trained CNN models, namely VGG-f~\cite{Chatfield-BMVC-14} and VGG-face~\cite{Parkhi-BMVC-2015}. We proceed by fine-tuning the networks in two stages. In the first stage (see details in Section~\ref{sec_train_full}), we fine-tune the CNN models on images with full faces. In the second stage (see details in Section~\ref{sec_train_half}), we further fine-tune the models on images in which the upper half of the face is occluded. 
All models are trained using data augmentation, which is based on including horizontally flipped images. To prevent overfitting, we employ Dense-Spare-Dense (DSD) training~\cite{Han-ICLR-2017} to train our CNN models. The training starts with a dense phase, in which the network is trained as usual. When switching to the sparse phase, the weights that have lower absolute values are replaced by zeros after every epoch. A sparsity threshold is used to determine the percentage of weights that are replaced by zeros. The DSD learning process, typically ends with a dense phase. It is important to note that DSD can be applied several times in order to achieve the desired performance. We next describe in detail each of the three training stages.

\subsection{Training on non-occluded faces}
\label{sec_train_full}

\noindent
{\bf VGG-face.}
VGG-face~\cite{Parkhi-BMVC-2015} is a deep neural network that is pre-trained on a closely related task, namely face recognition. The architecture is composed of $16$ layers. We keep its $13$ convolutional (conv) layers, replacing the fully-connected (fc) layers with a single max-pooling layer for faster inference. We also replace the softmax layer of $1000$ units with a softmax layer of $8$ units, since FER+~\cite{Barsoum-ICMI-2016} and AffectNet~\cite{Mollahosseini-TAC-2019} contain $8$ classes of emotion. We randomly initialize the weights in the softmax layer, using a Gaussian distribution with zero mean and $0.1$ standard deviation. We add $6$ dropout layers after each conv layer, starting from the fourth convolutional block. The first dropout layer has a dropout rate of $0.3$. For each subsequent dropout layer, the dropout rate increases by $0.05$. Thus, the last dropout layer has a dropout rate of $0.55$. We set the learning rate to $10^{-4}$ and we decrease it by a factor of $10$ when the validation error stagnates for more than $10$ epochs. In order to train VGG-face, we use stochastic gradient descent using mini-batches of $64$ images and set the momentum rate to $0.9$. We fine-tune VGG-face using DSD training~\cite{Han-ICLR-2017} for a total of $50$ epochs. For DSD training, we set the sparsity rate to $0.2$ for the second conv layer, increasing the rate up to $0.7$ with each additional conv layer. We refrain from pruning the weights of the first conv layer. Since VGG-face is pre-trained on a closely related task (face recognition), it converges in only $50$ epochs.

\noindent
{\bf VGG-f.}
We also fine-tune the VGG-f~\cite{Chatfield-BMVC-14} network with $8$ layers, which is pre-trained on the ILSVRC benchmark~\cite{Russakovsky2015}. As for VGG-face, we keep the conv layers of VGG-f, replacing the fc layers with a single max-pooling layer. We also replace the softmax layer of $1000$ units with a softmax layer of $8$ units. We add $3$ dropout layers after each conv layer, starting from the third convolutional layer. In each dropout layer, we set the dropout rate to $0.2$. We set the learning rate to $10^{-3}$ and we decrease it by a factor of $10$ when the validation error stagnates for more than $10$ epochs. At the end of the training process, the learning rate drops to $10^{-5}$. In order to train VGG-f, we use stochastic gradient descent using mini-batches of $512$ images and set the momentum rate to $0.9$. As for VGG-face, we use the DSD training method to fine-tune the VGG-f model. However, we refrain from pruning the weights of the first conv layer during the sparse phases. 
We set the sparsity rate to $0.2$ for the second conv layer, increasing the rate up to $0.5$ with each additional conv layer. We train this network for a total of $800$ epochs. Since VGG-f is pre-trained on a distantly related task (object recognition), it converges in a higher number of epochs than VGG-face.

\subsection{Training on occluded faces}
\label{sec_train_half}

\noindent
{\bf VGG-face.}
We fine-tune VGG-face on images containing only occluded faces, using, in large part, the same parameter choices as in the first training stage. The architecture is the same as in the first training stage. We hereby present only the different parameter choices. We applied DSD training for $40$ epochs, starting with a learning rate of $10^{-3}$, decreasing it by a factor of $10$ each time the validation error stagnated for more than $10$ epochs. At the end of the training process, the learning rate drops to $10^{-4}$.

\noindent
{\bf VGG-f.}
In a similar fashion, we fine-tune VGG-f on images containing only occluded faces, preserving the architecture and most of the parameter choices. We next describe the differences from the first training stage. We applied DSD training for $80$ epochs using a learning rate of $10^{-3}$. This time we did not have to decrease the learning rate during training, as the validation error drops after every epoch.

\section{Experiments}
\label{sec_experiments}

\subsection{Data sets}

\noindent
{\bf FER+.}
The FER+ data set~\cite{Barsoum-ICMI-2016} is a curated version of FER 2013~\cite{Goodfellow-ICONIP-2013} in which some of the original images are relabeled, while other images, e.g. not containing faces, are completely removed. Barsoum et al.~\cite{Barsoum-ICMI-2016} add \textit{contempt} as the eighth class of emotion along with the other $7$ classes: \textit{anger}, \textit{disgust}, \textit{fear}, \textit{happiness}, \textit{neutral}, \textit{sadness}, \textit{surprise}. The FER+ data set contains $25045$ training images, $3191$ validation images and another $3137$ test images. All images are of $48 \times 48$ pixels in size.

\noindent
{\bf AffectNet.}
The AffectNet~\cite{Mollahosseini-TAC-2019} data set contains $287651$ training images and $4000$ validation images, which are manually annotated. Since the test set is not publicly available, researchers~\cite{Mollahosseini-TAC-2019,Zeng-ECCV-2018} evaluate their approaches on the validation set containing $500$ images for each of the 8 emotion classes. 

\subsection{Implementation details}

The input images in both data sets are scaled to $224 \times 224$ pixels. We use the MatConvNet~\cite{matconvnet} library to train the CNN models. Each CNN architecture is trained on the joint AffectNet and FER+ training sets. On AffectNet, we adopt the down-sampling setting proposed in~\cite{Mollahosseini-TAC-2019}, which solves, to some extent, the imbalanced nature of the facial expression recognition task. As Mollahosseini et al.~\cite{Mollahosseini-TAC-2019}, we select at most $15000$ samples from each class for training. This leaves us with a training set of $88021$ images.

\subsection{Results}

\begin{table}[!t]
\caption{Accuracy rates of various VGG-f and VGG-face models on AffectNet~\cite{Mollahosseini-TAC-2019} and FER+~\cite{Barsoum-ICMI-2016}, using full faces or lower-half faces (occluded upper half) for training and testing. A bag-of-visual-words model~\cite{Ionescu-WREPL-2013} and two state-of-the-art CNN models (trained on full faces), VGG-13~\cite{Barsoum-ICMI-2016} and AlexNet~\cite{Mollahosseini-TAC-2019}, are also included for reference.}\label{tab_results}
\begin{center}
\begin{tabular}{|l|l|l|c|c|}
\hline
\bf Model       & \bf Train set      & \bf Test set     & \bf AffectNet     & \bf FER+\\
\hline
\hline
Bag-of-visual-words~\cite{Ionescu-WREPL-2013}  & full faces & full faces & $48.30\%$ & $80.65\%$\\
VGG-13~\cite{Barsoum-ICMI-2016} & full faces        & full faces & -        & $84.99\%$\\
AlexNet~\cite{Mollahosseini-TAC-2019} & full faces  & full faces & $58.00\%$ & - \\
\hline
VGG-f           & full faces        & full faces        & $57.37\%$         & $85.05\%$\\
VGG-face        & full faces        & full faces        & $59.03\%$         & $84.79\%$\\
\hline
VGG-f           & full faces        & lower-half faces  & $41.58\%$         & $70.00\%$\\
VGG-face        & full faces        & lower-half faces  & $37.70\%$         & $68.89\%$\\
\hline
VGG-f           & lower-half faces  & lower-half faces  & $47.58\%$         & $78.23\%$\\
VGG-face        & lower-half faces  & lower-half faces  & $49.23\%$         & $82.28\%$\\
\hline
\end{tabular}
\end{center}
\end{table}

In Table~\ref{tab_results}, we present the empirical results conducted on AffectNet~\cite{Mollahosseini-TAC-2019} and FER+~\cite{Barsoum-ICMI-2016} using either VGG-f or VGG-face. The models are trained and tested on various combinations of full or occluded images. We include a bag-of-visual-words baseline~\cite{Ionescu-WREPL-2013} and two state-of-the-art CNN models (trained on full faces), VGG-13~\cite{Barsoum-ICMI-2016} and AlexNet~\cite{Mollahosseini-TAC-2019}, for reference. First, we note that our models, VGG-f and VGG-face, attain results that are on par with the state-of-the-art CNNs, when full faces are used for training and testing. However, the CNN models trained on full faces give poor results when lower-half faces are used for testing. These results indicate that the CNN models (trained on full faces, as usual) are not particularly designed to handle severe facial occlusions, justifying our idea of fine-tuning the models on images containing such severe occlusions. Indeed, we observe significant accuracy improvements when VGG-f and VGG-face are fine-tuned on occluded images. For instance, the fine-tuning of VGG-face on lower-half images (occluded upper half) brings an improvement of $11.53\%$ (from $37.70\%$ to $49.23\%$) on AffectNet and an improvement of $13.39\%$ (from $68.89\%$ to $82.28\%$) on FER+. Interesting, our final models trained and tested on lower-half faces attain results that are not very far from the CNN models trained and tested on full faces. For example, our VGG-face model is $8.77\%$ under the state-of-the-art AlexNet~\cite{Mollahosseini-TAC-2019} on AffectNet, and $2.71\%$ under the state-of-the-art VGG-13~\cite{Barsoum-ICMI-2016} on FER+. Despite being tested on lower-half images, both VGG-f and VGG-face surpass the bag-of-visual-words model~\cite{Ionescu-WREPL-2013}, which is tested on full images. We thus conclude that our CNN models can provide reliable results, despite being tested on faces that are severely occluded.



\section{Conclusion}
\label{sec_conclusion}

In this paper, we proposed a learning approach based on fine-tuning CNN models in order to recognize facial expressions of severely occluded faces. The empirical results indicate that our learning framework can bring significant performance gains, leading to models that provide reliable results in practice, even surpassing a bag-of-visual-words baseline~\cite{Ionescu-WREPL-2013} tested on images with fully visible faces.


%
%
%
\bibliography{references}{} 
\bibliographystyle{splncs04}

\end{document}